\documentclass{article}
\usepackage{spconf,amsmath,graphicx}

\usepackage{amsmath}
\usepackage{algorithmic}
\usepackage{microtype}
\usepackage{graphicx}
\usepackage{pdfpages}
\usepackage{booktabs} % for professional tables
\usepackage{array}
\usepackage{url}
\usepackage{algorithm}
\usepackage{algorithmic}
\usepackage{bm}
\usepackage{amsmath}
\usepackage{amsfonts}
\usepackage{amssymb}
\usepackage{color}
\usepackage{subfigure}
\usepackage{cite}
\usepackage{balance}
\usepackage[breaklinks]{hyperref}

\def\z{{\mathbf z}}

\def\0{{\mathbf 0}}

\newcommand{\xc}{\bm{x}}

\newcommand{\zc}{\bm{z}}

\definecolor{cgray}{gray}{0.4}

\title{Continual Learning for Infinite Hierarchical \\ Change-Point Detection}
\name{Pablo Moreno-Mu\~noz, David Ram\'irez and Antonio Art\'es-Rodr\'iguez \thanks{This was supported by the Ministerio de Ciencia, Innovaci{\'o}n y Universidades under grant TEC2017-92552-EXP (aMBITION), by the Ministerio de Ciencia, Innovaci{\'o}n y Universidades, jointly with the European Commission (ERDF), under grants TEC2017-86921-C2-2-R (CAIMAN) and RTI2018-099655-B-I00 (CLARA), and by The Comunidad de Madrid under grant Y2018/TCS-4705 (PRACTICO-CM). The work of P. Moreno-Mu\~noz has also been supported by FPI grant BES-2016-077626.}}
\address{Dept. of Signal Theory and Communications, Universidad Carlos III de Madrid, Spain \\ Gregorio Mara\~n\'on Health Research Institute, Spain}
\begin{document}
\ninept
\maketitle
\begin{abstract}
Change-point detection (CPD) aims to locate abrupt transitions in the generative model of a sequence of observations. When Bayesian methods are considered, the standard practice is to infer the posterior distribution of the change-point locations. However, for complex models (high-dimensional or heterogeneous), it is not possible to perform reliable detection. To circumvent this problem, we propose to use a hierarchical model, which yields observations that belong to a lower-dimensional manifold. Concretely, we consider a latent-class model with an unbounded number of categories, which is based on the chinese-restaurant process (CRP).  For this model we derive a continual learning mechanism that is based on the sequential construction of the CRP and the expectation-maximization (EM) algorithm with a stochastic maximization step. Our results show that the proposed method is able to recursively infer the number of underlying latent classes and perform CPD in a reliable manner.
\end{abstract}
\section{Introduction}

Change-point detection (CPD), which consists of locating abrupt transitions in the generative model of the observations, is a problem with a plethora of applications. For instance, CPD is widely used in finance \cite{andersson2006some,berkes2004sequential}, the analysis of social networks \cite{raginsky2012sequential,krishnamurthy2012quickest}, or cognitive radio \cite{arts15,du15}. The main focus of CPD methods has been traditionally on batch settings, where the entire sequence of observations is available and has to be segmented. However, CPD is most useful in online scenarios, where change points must be detected as new incoming samples are observed. Online CPD methods have two intertwined tasks to solve: i) segmentation of sequential data into partitions (or segments) and ii) estimation of the generative model parameters for the given partitions. 

Since each partition has a different generative distribution, the identifiability of change points is related to the difference between such distributions. In this context, Bayesian inference is useful for inferring the distributions given a prior distribution in a reliable manner. The Bayesian online change-point detection (BOCPD) approach \cite{Adams2007} used this idea for recursively performing density estimation, which yields a more robust detection process as the propagation of uncertainty is considered. However, it can be observed that, for complex likelihood models, which have a number of parameters much higher than the number of observations between two consecutive change points, reliable CPD becomes unfeasible. This can be the case of, although is not restricted to, high-dimensional and/or heterogeneous observations (mixture of continuous and discrete variables), which usually have a prohibitive number of parameters.

To address the aforementioned issue, in \cite{MorenoRamirezArtes18} we presented a hierarchical probabilistic model based on latent classes, i.e., a mixture model. The CPD problem can be carried out directly on the lower-dimensional manifold, where the discrete latent variables lie. Hence, this method requires less evidence than the observational counterpart since the number of parameters is reduced, which yields faster and more reliable detections. However, \cite{MorenoRamirezArtes18} requires that the number of classes is fixed \emph{a priori}. 

The main contribution of this paper is to introduce a novel approach, based on continual learning \cite{Ring1994continual,Schmidhuber2013powerplay,Nguyen2018}, to recursively infer the underlying sequence of latent classes, its distributions, and the change points. The key idea of the proposed model is to allow for an unbounded order on the latent model, that is, the number of classes is not fixed and could even become infinite. In particular, we use the Chinese-restaurant process (CRP) \cite{Pitman2002}, which is a well-known Bayesian non-parametrics method, to model the latent variables with an unbounded number of classes. That is, the CRP may increase the number of classes as new observations come in. Moreover, as with any mixture model, the expectation-maximization (EM) algorithm \cite{EM_Dempster} is used, but in this work the maximization step (M-step) is substituted by a stochastic M-step \cite{Cappe2009}. Finally, the experimental results on real data show how both the latent-class inference process and the change-point detection perform reliably. 
%\vfill
%
\section{Bayesian Online Change-Point Detection}
\label{sec:bocpd}

We start by considering a time series $\xc_{1:t} = \{x_1, x_2, \ldots, x_t\}$, which is divided into non-overlapping partitions, denoted by $\rho_i, i = 1,2, \ldots$ Each partition is separated from its neighbors by change points (\textsc{cp}). Based on \cite{Adams2007}, we assume that the data within each partition $\rho_i$ is independent and identically distributed (i.i.d.) according to some generative probability distribution $p(x_t|\bm{\theta}_{\rho_i})$, where the parameter vector, $\bm{\theta}_{\rho_i}$, is unknown. Under this assumption, change points are determined by changes in the parameters:
\begin{equation}
\bm{\theta}_t = \begin{cases}
\bm{\theta}_{\rho_1}, & t < \textsc{cp}_1, \\
\bm{\theta}_{\rho_2}, & \textsc{cp}_1 \leq t \leq \textsc{cp}_2, \\
\bm{\theta}_{\rho_3}, & \textsc{cp}_2 \leq t \leq \textsc{cp}_3, \\
& \vdots
\end{cases}
\end{equation}
The main idea in \cite{Adams2007} is the \textit{run-length}, $r_t$, which is defined as a discrete random variable that counts the number of time-steps since the last \textsc{cp}, that is,
\begin{equation}
r_t = \begin{cases} 0, & \textsc{cp} \text{ at time } t \\
r_t + 1, & \text{otherwise}, \\
\end{cases}
\end{equation}
and may be seen as a proxy for change points. The objective of the BOCPD technique is to compute the posterior distribution $p(r_t|\xc_{1:t})$ recursively, from which we will identify a \textsc{cp} if the probability mass accumulates near $r_t=0$. 

\begin{figure}[t!]
	\centering
	\includegraphics[width=0.85\columnwidth]{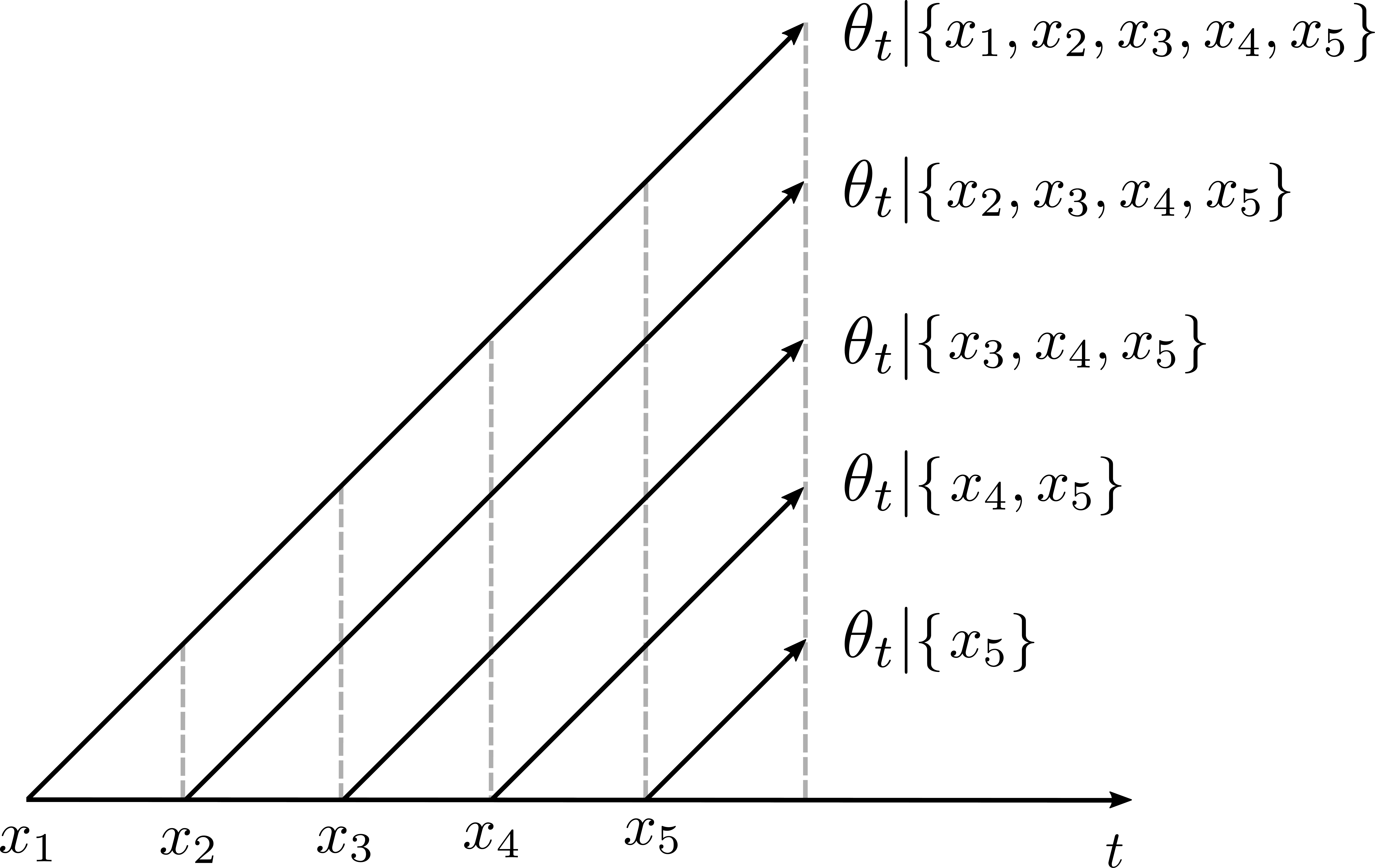}
	\caption{Illustration of the parallel inference threads for the estimation of $\bm{\theta}_t$ conditioned on the run-length $r_t$ given $\xc_{1:t}$.}
	\label{fig:thread}
\end{figure}

The posterior distribution $p(r_t|\xc_{1:t})$ is obtained by marginalizing the joint distribution $p(r_t,\xc_{1:t})$ over all the $r_t$ values seen so far, which, in turn, is computed by marginalizing the model parameters, $\bm{\theta}_t$. The learning of $\bm{\theta}_t$ given the partition, required for the computation of $p(r_t,\xc_{1:t})$, is carried out using a multiple thread inference mechanism induced by the run-length. For instance, to learn $\bm{\theta}_3$ given $r_3=2$, only the observations $\{x_2,x_3\}$ are required. This parallel inference scheme is depicted in Figure \ref{fig:thread}, where we illustrate the aforementioned example using the notation $\bm{\theta}_3 | \{x_2,x_3\}$.

The inference of $p(r_t|\xc_{1:t})$ in \cite{Adams2007} may become unfeasible when the complexity of the generative model increases, for instance, for high-dimensional and/or heterogenous observations.  That is, if the likelihood $p(x_t|\bm{\theta}_{\rho_i})$ for the partition $\rho_i$ depends on an extremely large number of parameters, it would not be possible to obtain sufficient statistical evidence to detect change points. This problem may yield the BOCPD method unusable in some problems.

% The particular details of the inference method are omitted here for the sake of space and because they are analogous to those presented in Section \ref{sec:bocpd_hierarchical}. Additionally, notice that depending on the likelihood model that is typically assumed to be an exponential family distribution, it would not be possible to obtain closed-form expressions.

\section{CPD on Hierarchical Models}
\label{sec:bocpd_hierarchical}

The aforementioned problem of the BOCPD for complex generative models can be overcome by introducing hierarchical models. We propose to use latent classes to obtain such hierarchical model. These latent classes, $z_t$, yield observations, $x_t$, that belong to a lower-dimensional manifold, and allow us to write the generative distribution of $x_t$ as
\begin{equation*}
p(x_t|\bm{\theta}_t) = \sum_{z_t = 1}^{K} p(x_t|z_t)p(z_t|\bm{\theta}_t), 
\end{equation*}
where $z_t$ is a categorical random variable, with $K$ being the maximum number of classes or categories, and $\bm{\theta}_t$ is the vector of parameters, i.e., the probability of each class. This form of latent-class model can be seen as a mixture model. 

Even assuming a hierarchical model, we are still interested in $p(r_t|\xc_{1:t})$, which would require the marginalization over $\zc_{1:t}$ as follows
\begin{equation}
 \label{eq:marginalization_z}
 p(r_t, \xc_{1:t}) = \sum_{\zc_{1:t}} p(r_t,\zc_{1:t}, \xc_{1:t}).
\end{equation}
However, for large values of $t$ and $K$, the marginalization in \eqref{eq:marginalization_z} is computationally unfeasible due to the combinatorial sums. In \cite{MorenoRamirezArtes18}, to avoid the marginalization, we assumed that we observe $\zc_{1:t}$, instead of marginalizing them, by directly plugging in the values of the \emph{maximum a posteriori} (MAP) estimates, which are given by
\begin{equation}
\label{eq:argmax}
 z_t^{\star} = \arg \max_{z_t} p(z_t|x_t).
\end{equation}
Now, using the MAP estimates as observations and assuming that the joint distribution on the right hand side (r.h.s.) of \eqref{eq:marginalization_z} factorizes as
\begin{equation*}
 p(r_t, \xc_{1:t},\zc_{1:t}^{\star}) = p(\xc_{1:t} | \zc_{1:t}^{\star}) p(r_t,\zc_{1:t}^{\star}),
\end{equation*}
with
\begin{equation*}
 p(r_t,\zc_{1:t}^{\star}) = \int p(r_t,\zc_{1:t}^{\star},\bm{\theta}_t) d \bm{\theta}_t,
\end{equation*}
we are effectively considering that the change points occurred on the sequence of latent classes. Using the extended recursion of \cite{MorenoRamirezArtes18}, which is given by 
\begin{equation}
\label{eq:recursive}
p(r_t, \zc_{1:t}^{\star}) %&= \sum_{r_{t-1}}p(r_t, r_{t-1}, \zc_{1:t}^{\star})\\
=\sum_{r_{t-1}} p(r_t|r_{t-1})\Psi^{(r)}_tp(r_{t-1}, \zc_{1:t-1}^{\star}),
\end{equation}
where $p(r_t|r_{t-1})$ is the conditional prior and
\begin{align}
 \Psi^{(r)}_t &= p(z_t^{\star}|r_{t-1}, \zc_{1:t-1}^{\star}) \nonumber \\
 &= \int p(z_t^{\star}|\bm{\theta}_t)p(\bm{\theta}_t|r_{t-1}, \zc_{1:t-1}^{\star}) d\bm{\theta}_t, \label{eq:predictive}
\end{align}
is the predictive distribution of the present latent variable conditioned on previous data and the run-length, we have all the ingredients to compute
\begin{equation}
\label{eq:posterior_runlength}
p(r_t | \zc_{1:t}^{\star}) = \frac{p(r_t, \zc_{1:t}^{\star})}{\sum_{r_t} p(r_t, \zc_{1:t}^{\star})},
\end{equation}
which determines the location of the change points.

\subsection{Infinite-dimensional Hierarchical BOCPD}
\label{sec:bocpd_hierarchical_infinite}

%\davidsays{Heeeere}

The problem of the hierarchical BOCPD algorithm presented above is that the number of classes, $K$, must be known and fixed \emph{a priori}. That is, $K$ is not allowed to vary over time, which can be a stringent condition in some scenarios. In this section, we consider the more interesting case that $K$ is unknown and can be time-varying, i.e., new classes may appear as $t\rightarrow \infty$. Then, we cannot select the order of the latent-class model in advance. A naive idea would be to fix an upper bound on $K$ and proceed as in the previous section. However, this upper bound could not be available and, even if it is, the performance can be poor, as we will see in Section \ref{sec:simulations}. In the following, we will present a method for unbounded and time-varying $K$, that is, $K$ is incremented when an unseen type of observations appears, which translates into a hierarchical BOCPD with unbounded $K$.

Using an unbounded number of classes  results in the following problem when integrating over $\bm{\theta}_t$ to compute $\Psi^{(r)}_t$. Assuming a Dirichlet distribution for $\bm{\theta}_t$, which is the conjugate prior for categorical distributions and therefore yields a tractable integral in \eqref{eq:predictive}, the evidence $p(z_t) \rightarrow 0$ as $K$ grows. To overcome this issue, we can consider an exchangeable distribution of the form $p([\zc_{t}]) = \sum_{\zc_{1:t} \in [\zc_{1:t}]}p(\zc_{1:t})$, where $[\zc_{1:t}]$ is a given division of classes, which is independent of the temporal assignments, i.e., $\zc_{1:3} = \{1,2,2\}$ corresponds to the same division of objects as $\zc_{1:3} = \{2,1,1\}$. This is often known as the \emph{exchangeability} property \cite{kingman1982,Pitman2002} and is a safe assumption in our setup as we are interested in changes in the probabilities of $\z_{t}$, not in the particular sequences $\zc_{1:t}$.

The latent-class model model with an unbounded dimension can be addressed using the Chinese-restaurant process (CRP) \cite{Pitman2002}, which is a Bayesian non-parametrics method \cite{orbanz2010}. The CRP is based on the metaphor where clients (observations $x_t$) are assigned to different tables (latent classes $z_t$) in a sequential manner. The assignment of classes to objects in the CRP is determined by the predictive posterior distribution, which is given by %\davidsays{I changed $m_{k,t}$ to $m_{k,t-1}$. Same for $K_t$}
\begin{equation}
\label{eq:crp_predictive}
p(z_t=k|z_1,\ldots,z_{t-1}) = \begin{cases} \frac{m_{k,t-1}}{t-1+\alpha}, & k\leq K_{t-1}, \\
\frac{\alpha}{t-1+\alpha}, & k = K_{t-1}+1, \\
\end{cases}
\end{equation}
where $m_{k,t-1}$ counts the number of assignments to class $k$ up to time $t-1$, $K_{t-1}$ is the number of classes associated with $m_{k,t-1}>1$ and $\alpha$ is a hyperparameter, which corresponds to the natural parameter of a symmetric Dirichlet prior distribution, and controls how likely is the appearance of a new class. 

Exploiting the aforementioned CRP construction, the computation of $\Psi^{(r)}_t$ in \eqref{eq:recursive} is straightforward, and is given by
\begin{equation}
  \label{eq:crp_predictive_MAP}
  \Psi^{(r)}_t = p(z^{\star}_t=k|r_{t-1},\zc^{\star}_{1:t-1}),
\end{equation}
%\begin{equation}
%  \label{eq:crp_predictive_MAP}
%  \Psi^{(r)}_t = \begin{cases} \frac{m_{k,t-1}^{(r)}}{t-1+\alpha}, & k\leq K_{t-1}, \\
%\frac{\alpha}{t-1+\alpha}, & k = K_{t-1}+1, \\
%\end{cases}
%\end{equation}
where we now count the number of MAP estimates, $z_t^{\star}$, equal to $k$ up to time $t-1$. Notice that this expression is analogous to \eqref{eq:crp_predictive} for a given run-length, i.e., for each parallel thread in Fig. \ref{fig:thread}. Then, we may proceed to compute the posterior $p(r_t| \zc_{1:t}^{\star})$.

One final comment is in order. So far, we have derived a tractable recursive way to introduce latent-class models into Bayesian CPD methods with an unbounded number of classes. However, nothing has been said on how to compute the MAP estimates in a continual learning fashion, which are required in \eqref{eq:posterior_runlength}. This task is explored in Section \ref{sec:continual}.

\section{Continual Learning of the CRP}
\label{sec:continual}

\begin{algorithm}[t!]
	\caption{Infinite-dimensional Hierarchical BOCPD}
	\label{alg:online_algorithm}
	\begin{algorithmic}[1]\small
		\STATE {\bfseries Input:} Observe $x_t$ and initialize $\hat{\bm{\varphi}}_{K_{t-1}}$.
		\STATE Sample $z_t \sim p(z_t|\zc_{1:t-1}^{\star})$ %\davidsays{Added $r_{t-1}$. Step 6?} 		
		\IF{$z_t  = K_{t-1} + 1$} \STATE Initialize {$\hat{\bm{\varphi}}_{K_{t-1}+1}$} \ENDIF
		\STATE Compute $p(z_t=k|\zc_{1:t-1}^{\star}), \forall k \leq K_{t-1}+1$
		\STATE Compute  $\mathbb{E}[\mathbb{I}\{z_t=k\}|\zc_{1:t-1}^{\star}, x_t, \hat{\bm{\varphi}}_k^{(t-1)}], \forall k \leq K_{t-1}+1$ %\comm{E-step}
		\STATE Update parameters $\{\hat{\bm{\varphi}}_k\}^{K_{t-1}+1}_{k=1}$ using \eqref{eq:m_step} %\comm{stochastic M-step}
		\STATE Calculate $z^{\star}_{t} = \arg\max(p(z_{t}|\zc_{1:t-1}^{\star}, x_t, \{\hat{\bm{\varphi}}_k^{(t)}\}_{k = 1}^{K_{t-1}+1})$ %\comm{MAP}
		\IF{$z^{\star}_{t} = K_{t-1} + 1$} \STATE {$K_t = K_{t-1} + 1$} \ENDIF
		\FOR{$r_t=1$ {\bfseries to} $t$}
		\STATE Evaluate $\Psi^{(r)}_t$ using \eqref{eq:crp_predictive_MAP} % \comm{CRP predictive}
		\STATE Calculate $p(r_{t},\zc_{1:t}^{\star})$ 
		\STATE Obtain  $p(\zc_{1:t}^{\star}) = \sum_{r_{t}}p(r_{t},\zc_{1:t}^{\star})$ % \comm{marginalization}
		\STATE Compute $p(r_{t}|\zc_{1:t}^{\star})$ 
		\STATE Update $m^{(r)}_{k,t} \leftarrow m^{(r)}_{k,t-1} + \mathbb{I}\{z_{t}^{\star} = k\}$ % \comm{counter}
		\ENDFOR
		\STATE{\bfseries Return:} $r^{\star}_{t}=\arg\max p(r_{t}|\zc_{1:t}^{\star})$ %\comm{change-point}
	\end{algorithmic}
\end{algorithm}

%\davidsays{You did not include the comment about incrementing $K_t$ by $1$, and only keep it if the MAP estimate is $z_t^{\star} = K_t + 1$.}

In this section, we compute the MAP estimates of $z_t$ in an online and recursive fashion. This task also involves the estimation of $\{\bm{\varphi}_k\}_{k = 1}^{K_t}$, which are the parameters of the mapping between observations and latent variables, that is, $p(x_t | z_t = k) = p(x_t | z_t = k, \bm{\varphi}_k)$. Here, the number of classes $K_{t}$ increases if when sampling from the CRP predictive distribution the result is $K_{t-1} + 1$. That is, at the beginning of each iteration we create a new class with an emission probability given by \eqref{eq:crp_predictive}, which is only kept if the MAP estimate is $z_t^{\star} = K_{t-1} + 1$. %\davidsays{I've added this, but it might be a bit confusing.}

Mixture models do not usually have closed-form solutions for the estimates of the parameters and the class assignments. Therefore, it is necessary to resort to the expectation-maximization (EM) algorithm \cite{EM_Dempster}, for which we need the log-likelihood of the complete data, which is given by
\begin{multline}
\label{eq:log_lik}
\mathcal{L}_{\bm{\varphi}}(\xc_{1:t},\zc_{1:t}) = \log p(\xc_{1:t},\zc_{1:t}|\{\bm{\varphi}_k\}^{K_t}_{k=1}) = \\
\log p(\zc_{1:t}) + \sum_{\tau = 1}^{t} \log p(x_{\tau}|z_{\tau}, \{\bm{\varphi}_k\}^{K_t}_{k=1}),
\end{multline}
where the prior distribution $p(\zc_{1:t})$ factorizes as 
\begin{equation*}
	p(\zc_{1:t}) = p(z_t|\zc_{1:t-1})p(z_{t-1}|\zc_{1:t-2})\cdots p(z_1).
\end{equation*}
This factorization is possible due to the chain-rule and the CRP construction described in Section \ref{sec:bocpd_hierarchical_infinite}. Once the complete data log-likelihood is available, we may apply the expectation step (E-step) and the maximization step (M-step) of the EM algorithm. In this work, we have slightly modified the M-step to accept the proposed continual learning framework. concretely, the estimation of the parameter at each step is simply performed by taking one iterate of a steepest descent method, yielding a stochastic M-step \cite{Cappe2009}. The E-step amounts to %\davidsays{Run-length?}
\begin{multline*}
	\mathbb{E}[\mathbb{I}\{z_t=k\}|\zc_{1:t-1}^{\star}, x_t, \hat{\bm{\varphi}}_k^{(t-1)}] =p(z_t=k|\zc_{1:t-1}^{\star}, x_t, \hat{\bm{\varphi}}_k^{(t-1)}) 
	\\ \propto p(x_t|z_t = k, \hat{\bm{\varphi}}_k^{(t-1)})p(z_t=k|\zc_{1:t-1}^{\star}),
\end{multline*}
where $\mathbb{E}[\cdot]$ is the expectation operator, $\hat{\bm{\varphi}}^{(t)}_k$ is the estimate of $\bm{\varphi}_k$ at time $t$, and we have exploited \eqref{eq:crp_predictive}. In the M-step, the estimate of the parameters $\{\bm{\varphi}_k\}^{K_t}_{k=1}$ is updated based on the gradient:
\begin{equation}
\label{eq:m_step}
\hat{\bm{\varphi}}^{(t)}_k \leftarrow \hat{\bm{\varphi}}^{(t-1)}_k + \eta_{k,t} \nabla_{\bm{\varphi}_k} \mathbb{E}[\mathcal{L}_{\bm{\varphi}}(\xc_{1:t},\zc_{1:t})],
\end{equation}
where $\eta_{k,t}$ is the (adaptive) learning rate for the $k$th class at time $t$. In this expression, we have assumed that the same initial learning rate is chosen for the parameters of a given class, but it is possible to select multiple learning rates per class. Once we have the E- and M-steps, we can compute the posterior of $z_t$ and maximize it to obtain $z_t^{\star}$ as in \eqref{eq:argmax}. Finally, Algorithm \ref{alg:online_algorithm} presents all the necessary computations of the proposed recursive method at each time instant $t$ and the Python implementation can be found in \small{\url{https://github.com/pmorenoz/continual_ihcpd}} for reproducibility purposes. \normalsize

\section{Experiments}
\label{sec:simulations}

\begin{figure*}[t!]
	\centering
	\includegraphics[width=0.48\textwidth]{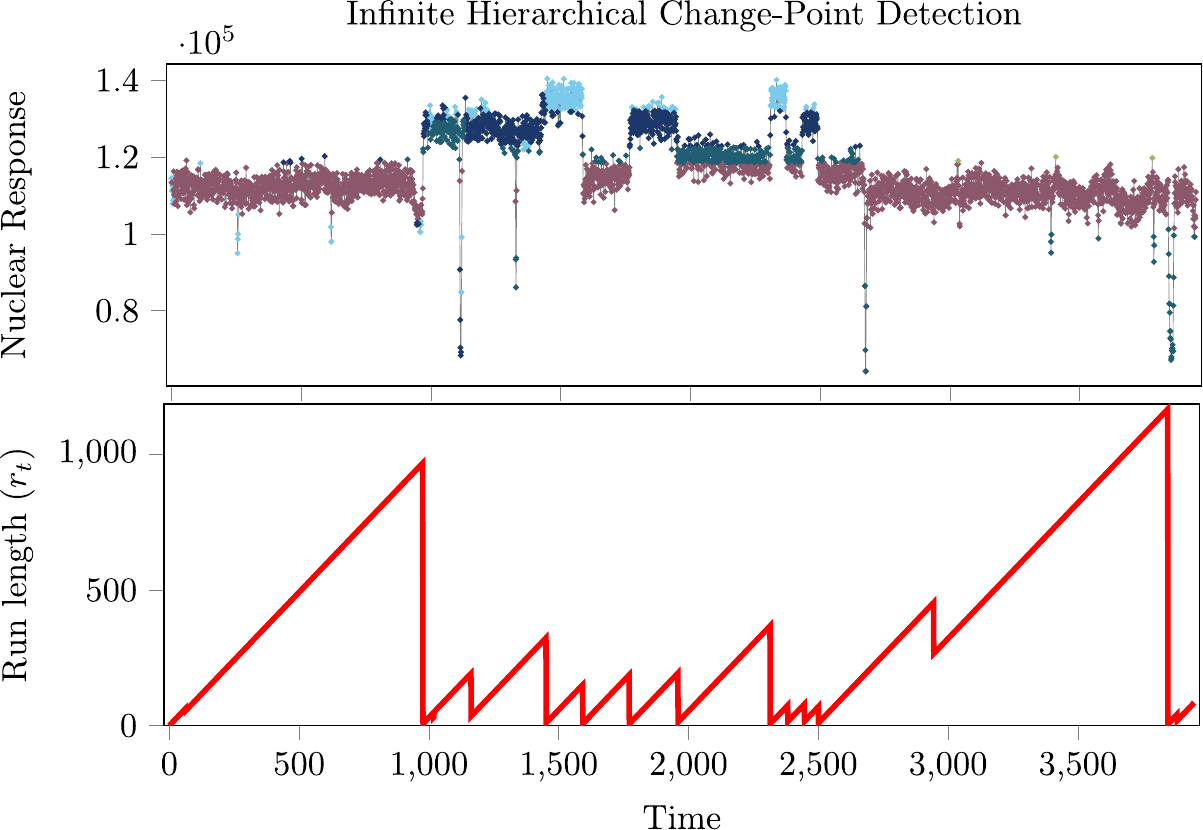}
	\includegraphics[width=0.48\textwidth]{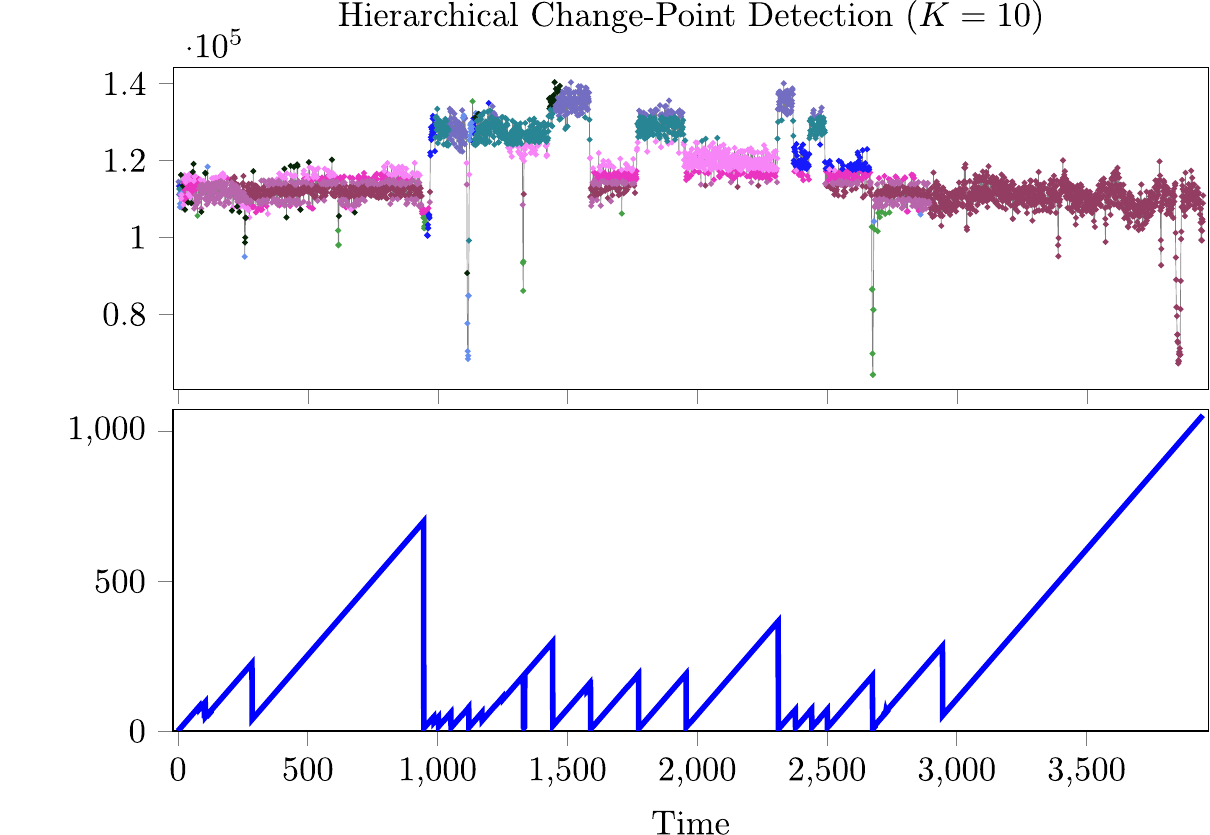}
	\caption{Upper row plots show the well-drilling univariate signal for the unbounded latent variable model (left) and the hierarchical CPD method (right) with fixed $K$. The colors represent latent-class asignments. Bottom row plots show the MAP estimates of the run-length.}
	\label{fig:continual}
\end{figure*}

In this section we evaluate the performance of the proposed method. We apply the infinite-dimensional hierarchical BOCPD algorithm to real-world data, and in particular, to a sequence of raw nuclear magnetic response measurements taken during a well-drilling process. This data consists of $4500$ real-valued univariate observations taken at a fixed sampling frequency. In the following, we assume that the time steps are ordered and discrete for simplicity. 

To apply the proposed model, we choose $p(x_t|z_t = k, \bm{\varphi}_k)$ to be Gaussian distributed with unknown mean and variance, that is, $\bm{\varphi}_k = \{\mu_k, \sigma_k^2\}$. Moreover, the model has two hyperparameters that we need to select. The first one, which is related to the CPD method, is the parameter $\lambda$ of the hazard function that is used as the conditional prior, $p(r_t|r_{t-1})$. In the experiments, we have selected $\lambda = 10^6$. The second one is the parameter $\alpha$, which is involved in the CRP construction, and controls how likely is the appearance of a new unseen class. We set it to $\alpha = 1.0$. For the stochastic M-step, we use two different adaptive learning rates for the mean and variance whose initial values are given by $\eta_{\mu} = 1.0$ and $\eta_{\sigma} = 0.02$. Importantly, we made both learning rates decrease at a rate of $2\%$ per time-step if $z_t=k$ was selected as the most likely latent class. This choice avoids adapting very old parameters with new incoming data.

Figure \ref{fig:continual} shows the results obtained for $t=4500$ iterations.\footnote{A video demonstrating the complete simulation of the algorithms is available at \url{https://www.youtube.com/watch?v=ymZPNURhtIc}.} The unbounded model is compared with the hierarchical CPD approach with an upper bound on the number of classes $K = 10$. In the upper figures we can see the well-drilling signals, as well as the latent-class assignments in different colors for both approaches. As can be observed, the final number of classes inferred by the CRP was $K_{4500}=7$. In the bottom figures we show the MAP estimates of the run-length, $r_t^{\star}$. These figures show that the MAP estimation of the run-length aligns quite well with the signal transitions. Furthermore, it should be noted that the proposed method is more robust to outliers as can be seen for $t \approx 200$ and $t \approx 600$, where the outlier is captured by the latent class assignment but a CP is not declared. In fact, the MAP estimate of the run-length is noisier for the method with a fixed number of classes than for the unbounded model. 

Finally, it is important to note that, since the unbounded model creates new classes as they become necessary, its computational complexity is smaller than that of hierarchical CPD approach, which needs to estimate the parameters of $K = 10$ classes at every time step.

\section{Discussion and Future Work}

%\davidsays{I do not like this conclusion.}

This work has extended the Bayesian online change-point detection method to more complex scenarios by considering a hierarchical model, which is based on latent-class variables. To prevent the limitation of fixing the order of the hierarchical model \emph{a priori}, we allow for an unbounded number of classes using the chinese restaurant process. Moreover, the inference of the class assignments is done with an expectation-maximization algorithm, where the M-step is carried out stochastically, that is, only one iteration of a steepest descent method is taken. Finally, the performance of the proposed method is validated empirically over real-world data. We show its robustness and utility for the aforementioned purposes. In future work, it would be interesting to extend it to multi-channel settings with multivariate generative models.

\balance
\bibliographystyle{IEEEtran}
\bibliography{IEEEabrv,paper_icassp}

\end{document}